\RequirePackage{fix-cm}
\documentclass[twocolumn]{svjour3}          
\smartqed  
\usepackage{graphicx}
\usepackage[authoryear]{natbib}
\usepackage{amsmath}
\usepackage{float}
\usepackage{subfigure}
\usepackage{algorithm}
\usepackage{algpseudocode}
\usepackage{multirow}
\usepackage{booktabs}
\usepackage{url}
\algnewcommand{\algorithmicglobal}{\textbf{Global:}} 
\algnewcommand\Global{\item[\algorithmicglobal]}

\journalname{Journal of Scheduling}
\begin{document}

\title{Neural Networked Assisted Tree Search for the Personnel Rostering Problem\thanks{This work was supported by the National Natural Science Foundation of China [71901214; 71690233]; the AI in Flanders, KU Leuven Research Fund [RKU-D2932-C24/17/012]; and the Data-driven logistics [FWO-S007318N].}
}


\author{Ziyi Chen         \and
	Patrick De Causmaecker \and
	Yajie Dou
}


\institute{Ziyi Chen \at
	College of Systems Engineering, National University of Defense Technology, Changsha 410073, China \\
	Tel.: +32-485924591\\
	\email{ziyi.chen@student.kuleuven.be}\\
	\emph{Present address:Department of Computer Science, KU Leuven Kulak, E. Sabbelaan 53, 8500 Kortrijk, Belgium}    
	\and
	Patrick De Causmaecker \at
	Department of Computer Science, KU Leuven Kulak, E. Sabbelaan 53, 8500 Kortrijk, Belgium\\
	\and
	Yajie Dou \at
	College of Systems Engineering, National University of Defense Technology, Changsha 410073, China
}

\date{Received: date / Accepted: date}

\maketitle

\begin{abstract}
The personnel rostering problem is the problem of finding an optimal way to assign employees to shifts, subject to a set of hard constraints which all valid solutions must follow, and a set of soft constraints which define the relative quality of valid solutions. The problem has received significant attention in the literature and is addressed by a large number of exact and metaheuristic methods. In order to make the complex and costly design of heuristics for the personnel rostering problem automatic, we propose a new method combined Deep Neural Network and Tree Search. By treating schedules as matrices, the neural network can predict the distance between the current solution and the optimal solution. It can select solution strategies by analyzing existing (near-)optimal solutions to personnel rostering problem instances. Combined with branch and bound, the network can give every node a probability which indicates the distance between it and the optimal one, so that a well-informed choice can be made on which branch to choose next and to prune the search tree.
\end{abstract}

\section{Introduction}
In various occupations and work scenarios, arranging employees to different shifts is a difficult job. The difficulty is that different employees have different requirements for life and work, which leads to preference of each employee. And there are also  requirements of the law that must be followed or diverse properties of different occupations. These regulations are what we call soft constraints and hard constraints. Inflexible or unreasonable work schedules may affect the personal lives of employees, affect their emotion, make them dissatisfied with their work, and may lead to a high turnover rate, which may have an adverse effect on the employer's operation and the experience impact of customers. If it can ensure that employees are arranged for the right job at the right time. From the perspective of employees, such schedule can improve employee job satisfaction, reduce employee dissatisfaction, fatigue and pressure to improve work efficiency and service quality\citep*{Burke2004}. From the perspective of employers and companies, an excellent schedule can increase the retention rate of employees and maintain a reasonable financial budget for employers.\citep*{kazahaya2005harnessing,MHallah2013}\par

Committed to solving such real-world problems, the personnel rostering problems have received great attention in the past decade. The personnel rostering problems aim to generate a scheduling table based on the determined number of employees and the time period. The schedule consists of a series of different types of shifts (for example, morning, evening, and day-off) during the entire time period. The basis for shifts is based on conditions such as the preference of employees for working hours and the requirements of laws or professional regulations, which we call hard constraints and soft constraints. Hard constraints are conditions that must be met for shift scheduling, while soft constraints can be violated to a certain extent, but must pay a price for this. The quality of the schedule can be evaluated by the penalty value for violating soft constraints. Due to the complex and highly-constrained structure, personnel rostering problems are often computationally challenging, and most variants of these problems are classified as NP-hard.\par 

There has been a lot of research on personal rostering problems, which can be divided into two categories: exact methods and metaheuristic methods.\citep*{Smet2016} The exact methods mainly include Integer Programming (IP)\citep*{Glass2010,Maenhout2009,MHallah2013} and Constrained Programming (CP) \citep*{girbea2011constraint,soto2013nurse}, the exact method can find the optimal solution, but the time cost it pays is also very expensive, and it is usually unacceptable. To solve this problem, the researchers have proposed metaheuristic methods, including Variable Neighborhood Search\citep*{Lue2012,Rahimian2017}, Genetic Algorithms\citep*{Ayob2013,Burke2001} and stochastic algorithms \citep{tassopoulos2015alpha} and tailored heuristic algorithms \citep*{brucker2010shift}, these methods can generate high-quality feasible solution in a short time, but the shortened time comes at the cost of giving up accuracy.\par

And even so, these complex methods are not well applied. According to the literature, many organizations are still manually scheduling schedules. The research has contributed to producing solutions automatically.\citep*{Burke2003,Causmaecker2010} \par

We propose a combined Deep Neural Network and Tree Search(DNNTS)based methdodology. DNNTS is the first method that used Deep Neural Network(DNN) model to guide tree search to solve the personnel rostering problem. It is based on a learned model for branch selection during the tree search. The implementation can be used to solve a number of problems from the literature\citep{Schedulingbenchmarks}. These problems have the following characteristics and observations:

\begin{enumerate}
	\item The (intermediate) solutions of a problem can be represented as m×n matrices that can be transferred as input for the neural network.
	\item The best solution can be found by modifying the initial solution in a number of simple steps which defining the possible child nodes.
	\item The problems have soft constraints which can be expressed in the penalty function and set the lower bound for the tree search.
\end{enumerate}

The main contributions of this work can be summarized as follows: 
\begin{enumerate}
	\item An automatic method DNNTS is applied to personnel rostering problem.
	\item For some specific problems, this method is shown to find a good solution equal to the best known lower bound.
	\item The Deep Neural Network is used to make branch selection in the process of tree search, which speed up the search process.
	\item Experimental evaluation of different search strategies.
\end{enumerate}

This paper is organized as follows. In Chapter 2, we discuss work on personal rostering and on combining machine learning and optimization techniques. In Chapter 3, we describe the problem and formal problem definition. In Chapter 4, we introduce the tree search method, the DNN model and the process of combining these to solve the personal rostering problem. In Chapter 5 we test our method, show the results and make relevant comparisons. In Chapter 6 we summarize and introduce the future work.\par
\section{Literature review}

In this chapter, we first review existing methods for personal rostering. Similar methods combining deep learning and optimization are discussed. There are many optimization methods that integrate deep learning methods in other fields. We summarize the methods that inspired us, and discuss the connection between these and our methods. 

\subsection{Rostering problem}

As mentioned before, the methods to solve personal rostering problems are mainly divided into two categories, exact methods and heuristic methods. Exact method are mainly concerned with finding proven optimal solutions. For the direction of IP, \citet{Maenhout2009} present an exact branch-and-price algorithm for solving the nurse scheduling problem incorporating multiple objectives and discuss different branching and pruning strategies. Some authors\citep{girbea2011constraint,soto2013nurse} concentrate on CP, and introduce a model including soft constraints. Mixed integer programming(MIP) is also a method that has attracted much attention. \citet{MHallah2013} describe the nurses’ timetabling problem of a Kuwaiti health care unit and models it as a MIP. \citet{Glass2010} start from benchmark problems and extend their MIP approach to nurse rostering to take better account of the practical considerations. Column generation has been effective to determine preference scheduling \citep*{Bard2005}. applied Mixed Integer Quadratic Programming.\par
Heuristic methods are mainly concerned with finding a solution quickly, and even if the number of employees waiting to be scheduled is large and the constraints are complicated, an acceptable and feasible solution can be obtained in an appropriate time. Genetic and Memetic algorithms form an important class of metaheuristics that have been extensively applied in personnel rostering problem.\citep*{aickelin2000exploiting,Aickelin2004,Easton1999,934317}\citep{Burke2001} In addition, there are a lot of attempts on other types of methods, such as Tabu Search Algorithms, some researchers \citep*{10.1007/3-540-48873-1_25} proposed a hybrid Tabu Search Algorithm to solve the personnel rostering problem in Belgian hospitals. Also simulated annealing algorithms. As a representative, an iteratively local searching method based on simulated annealing\citep*{4811530}, and a shift mode method using simulated annealing were proposed.\citep*{5561304} \citet{Aickelin2004} proposed a mode conversion technique arising at a major UK hospital. And \citet*{todorovic2013bee} proposed the Bee Colony Optimization method to solve this problem.\par
\citet{Burke2004} analyze the state of the art of research on nurse rostering problems and categorized papers according to solution methods, constraints, performance measures, and information on the planning period, the data that is used, the number of skills, and their substitutability, etc. \citet{Causmaecker2010} build on the work of the last decades to produce a $\alpha$$|\beta|$$\gamma$ classification system for nurse rostering problems. \citet*{Bergh2013} present a review of literature on personnel scheduling and evaluate the literature from different perspectives of personnel characteristics, decision delineation and shifts definitions, constraints, performance measures, flexibility, application area and applicability of research.

\subsection{Deep learning and Optimization}
For the combination of deep learning and optimization, some researchers have made some attempts. For example, \citet*{Hottung2020} use two DNN models to guide the tree search to solve Containers Pre-Marshalling Problem. As is mentioned in the paper written by \citet*{10.1007/978-3-319-67308-0_2}, data science meets optimization when using data science in algorithm construction and applying deep learning while engineering an algorithm. \par
Combining operations research and artificial intelligence allows using powerful solvers such as IBM CP Optimiser\citep*{optimizer2015v12} and Gurobi \citep*{gurobi2015gurobi}, to be used in hybrid environments. e.g. a method combing CP an heuristics, which is an iterated local search framework that uses CP for initial solution construction and diversification, and a variable neighborhood descent for iterative improvement. \citep*{10.1007/978-3-642-23786-7_9}. Another method divides the original problem into sub-problems, solves sub-problems by IP, and combines IP and local search to get results \citep*{Valouxis2012}. And a less studied combination of IP and CP \citep*{rahimian2015hybrid} allow to take advantage of the complementarity in the different methodologies. \citet{rahimian2015hybrid} proposed a new hybrid algorithm of IP and CP to solve the personnel rostering problem. It uses IP to find the best solution, and CP to find feasible solutions effectively. This hybrid algorithms uses information from specific problems to reduce the search space, fine-tunes search parameters and improves the efficiency of the entire search process in a novel way. \par
\citet*{Lodi2017} and \citet*{dilkina2017comments} outline methods for applying learning to variable and node selection problems in MIP. \citet*{khalil2017learning} use logistic regression to predict when to apply the original heuristic when solving MIPs. The author uses similar features as mentioned by \citet*{khalil2016learning} and can improve the performance of the MIP solver. There are other papers that use machine learning techniques to solve MIP \citep*{kruber2017learning,bonfietti2015embedding}.\par
\citet*{vinyals2015pointer} propose a method called a pointer network and train it to generate solutions to traveling salesman problems through supervised learning. \citet*{bello2016neural} use reinforcement learning to train a pointer network for the traveling salesman problem. \citet*{kool2018attention} propose a similar method, which can also be used to solve other routing problems, such as vehicle routing problems. All these methods focus on the training and architecture of the DNNs network, rather than merging the DNNs network into a complex search process.\par

\section{Problem description}

There are many personnel rostering problems in different work contexts, such as nurse scheduling, hotel reception scheduling or other situations. In the personnel rostering problems, the people waiting to be assigned are called employees, the different time in everyday waiting to be occupied are called shifts. Employees are assigned to shifts in a certain period of time according to certain constraints. Constraints define the limitations of assignments for each employee, there are hard constraints which the solution must obey and soft constraints which will give some penalty when the solution does not meet the requirements. These constraints can be used to model restrictions such as ‘employees should work more than 3 days and less than 6 days a week’ or ‘employee A does not want to work on Wednesday’\citep*{Causmaecker2010,paul2015classification}. The obtained feasible assignments that meet all hard constraints are called solutions. Each solution has a value of penalty, which is determined by the level of compliance with soft constraints. The personnel rostering problem aims to find an allocation scheme with the lowest penalty value that meets all hard constraints. It is an NP-hard problem\citep*{osogami2000classification}.\par 

\subsection{Formal problem definition}
\begin{figure*}
	\includegraphics[width=1\textwidth]{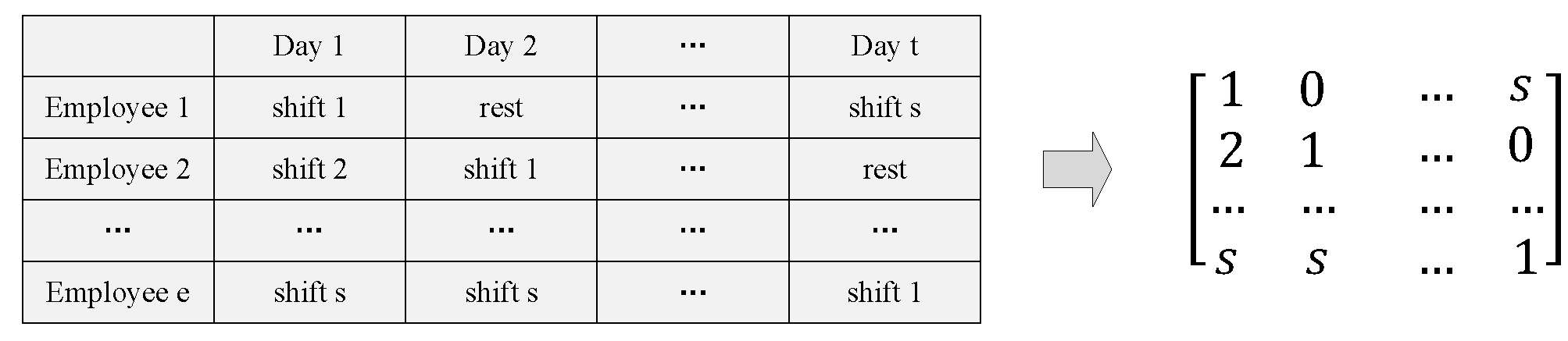}
	\caption{Solution and corresponding mathematical matrix expression}
	\label{Fig2}       
\end{figure*}

Personnel rostering problems are characterized by a set of employees $E=\{1,2,…,e\}$, a scheduling period of days $T=\{1,2,…t\}$, and a set of shifts $S=\{1,2,…,s\}$.A shift is a fixed time interval which denotes a working period. Each shift is characterized by a unique type which classifies the shifts in various ways, e.g., by time interval (morning, late), by required qualifications (senior, junior), or by a combination of these (morning-senior, late-junior). A shift is considered to occur on the day where its time interval starts. The number of employees required for each shift can vary from day to day, and is typically more than one employee. An assignment is the allocation of an employee to a shift on a day. In this paper, the solution is regarded as an e×t matrix which contains in each cell either an assignment or a day-off. Days T is a period during which the assignment begins and ends. Fig.\ref{Fig2} shows how to transfer a solution to a matrix.\par

Constraints can be expressed as an exact, ranged, minimum or maximum requirement. In the case of exact demand, the specified value is exactly the number of employees to be assigned. A ranged definition requires that the number of assigned employees should be within a specified time interval. When such an interval has no upper(lower) limit, the requirement is defined as a minimum(maximum).\citep{Smet2016}\par

Let $E$, $T$, $S$,$W$ be the set of employees, a scheduling period of days, a set of shifts, and set of weekends in the planning horizon. Let $h$ be the number of days in the planning horizon. Let $m_{es}^{max}$ be the maximum number of shifts of type $s$ that can be assigned to employee $e$. Let $b_e^{min}$,$b_e^{max}$ be the minimum and maximum number of minutes that employee $e$ must be assigned. Let $q_{ets}$, $p_{ets}$ be the penalty if shift type $s$ is not assigned and assigned to employee $e$ on day $t$. Let $v_{ts}^{min}$, $v_{ts}^{max}$ be the weight if below and exceeding the preferred cover for shift type $s$ on day $t$. Let $y_{ts}$, $z_{ts}$ be the total below and above the preferred cover for shift type $s$ on day $t$. Let $c_e^{min}$,$c_e^{max}$ be the minimum and maximum number of consecutive shifts that employee $e$ must work. Let $o^{min}_e$ be the number of consecutive days off that employee $e$ can be assigned. Let $a^{max}_e$ be the maximum number of weekends that employee $e$ can work.Then the problem can be formulated below:\par

\[
x_{ets}=\left\{
\begin{aligned}
&1& if\enspace employee\enspace e\enspace is\enspace assigned\enspace shift\enspace type\enspace s\enspace on\enspace day\enspace t \\
&0& otherwise 
\end{aligned}
\right.\]

\[
k_{ew}=\left\{
\begin{aligned}
&1& if\enspace employee\enspace e\enspace works\enspace in\enspace weekend\enspace w\\
&0& otherwise 
\end{aligned}
\right.\]

\begin{equation}
\begin{split}\label{equ1}
\min&\sum_{e\in E}\sum_{t\in T}\sum_{s\in S}{q_{ets}(1-x_{ets})}+\sum_{e\in E}\sum_{t\in T}\sum_{s\in S}{p_{ets}x_{ets}}+\sum_{t\in T}\sum_{s\in S}{y_{ts}v_{ts}^{min}}\\
&+\sum_{t\in T}\sum_{s\in S}{z_{ts}v_{ts}^{max}}
\end{split}
\end{equation}

\begin{equation}
\begin{split}\label{equ2}
\sum_{s\in S}{x_{ets}\leq 1},\quad \forall e\in E,t\in T
\end{split}
\end{equation}

\begin{equation}
\begin{split}\label{equ3}
x_{ets}+x_{e(t+1)s^{\prime}}\leq 1,\quad \forall e\in E,t\in \left\{1,...,h-1\right\},s,s^{\prime}\in S
\end{split}
\end{equation}

\begin{equation}
\begin{split}\label{equ4}
\sum_{t\in T}{x_{ets}\leq m_{es}^{max}},\quad \forall e\in E,s\in S
\end{split}
\end{equation}

\begin{equation}
\begin{split}\label{equ5}
b_e^{min}\leq \sum_{t\in T}\sum_{s\in S}{l_s x_{ets}}\leq b_e^{max},\quad
\forall e \in E
\end{split}
\end{equation}

\begin{equation}
\begin{split}\label{equ6}
\sum_{j=d}^{d+c_e^{max}}\sum_{s\in S}{x_{ejs}}\leq c_e^{max}, \quad \forall e\in E, d\in \left\{1...h-c_e^{max} \right\}
\end{split}
\end{equation}

\begin{equation}
\begin{split}\label{equ7}
\sum_{s\in S}{x_{ets}}&+(k-\sum_{j=t+1}^{t+k}\sum_{s\in S}{x_{ejs}}+\sum_{s\in S}{x_{i(t+k+1)s}}>0,\\
&\quad \forall e\in E, k\in \left\{1...c^{min}_e-1\right\},t\in \left\{1...h-(k+1)\right\}
\end{split}
\end{equation}

\begin{equation}
\begin{split}\label{equ8}
(1-\sum_{s\in S}&{x_{ets}})+\sum_{j=t+1}^{t+k}\sum_{s\in S}x_{ejs}+(1-\sum_{s\in S}{x_{i(t+k+1)s}})>0
\\
&\quad \forall e\in E, k\in \left\{1...o^{min}_e-1\right\},d\in \left\{1...h-(k+1)\right\}
\end{split}
\end{equation}

\begin{equation}
\begin{split}\label{equ9}
\sum_{w\in W}{k_{ew}}\leq a_e^{max}\quad \forall e\in E
\end{split}
\end{equation}

\begin{equation}
\begin{split}\label{equ10}
x_{ets}=0, \quad \exists e\in E,t\in T,s\in S
\end{split}
\end{equation}

Equation(\ref{equ1}) is the objective function. Constraint(\ref{equ2}) ensures that an employee cannot be assigned more than one shift on a single day. Constraint(\ref{equ3}) describes some types of shifts cannot follow others. Constraint(\ref{equ4}) is used to limit the maximum number of shifts of each type that can be assigned to employees. For example, some employees will have contracts which do not allow them to work night shifts or only a maximum number of night shifts. Constraint(\ref{equ5}) ensures that the minimum and the maximum work time. The total minutes worked by each employee must be between a minimum and maximum. These limits can vary depending on whether the employee is full-time or part-time. Constraints(\ref{equ6}) and (\ref{equ7}) describe the consecutive constraints. Constraint(\ref{equ8}) models the minimum consecutive days off in a similar way to the minimum consecutive shifts constraint. Constraint(\ref{equ9} sets the maximum number of weekends. A weekend is considered as being worked if the employee has a shift on the Saturday or the Sunday. Constraint(\ref{equ10} describes some days that employees cannot work, for example, they are on vacation.\citep*{curtois2014computational}\par

\section{Method}
The DNNTS method integrates DNN into a heuristic tree search to decide which branch to choose next. The DNN is trained offline by supervised learning on existing (near-) optimal solutions for the defined personnel rostering problem and are then used to make branch decisions during the search.\par
\begin{figure*}
  \includegraphics[width=1\textwidth]{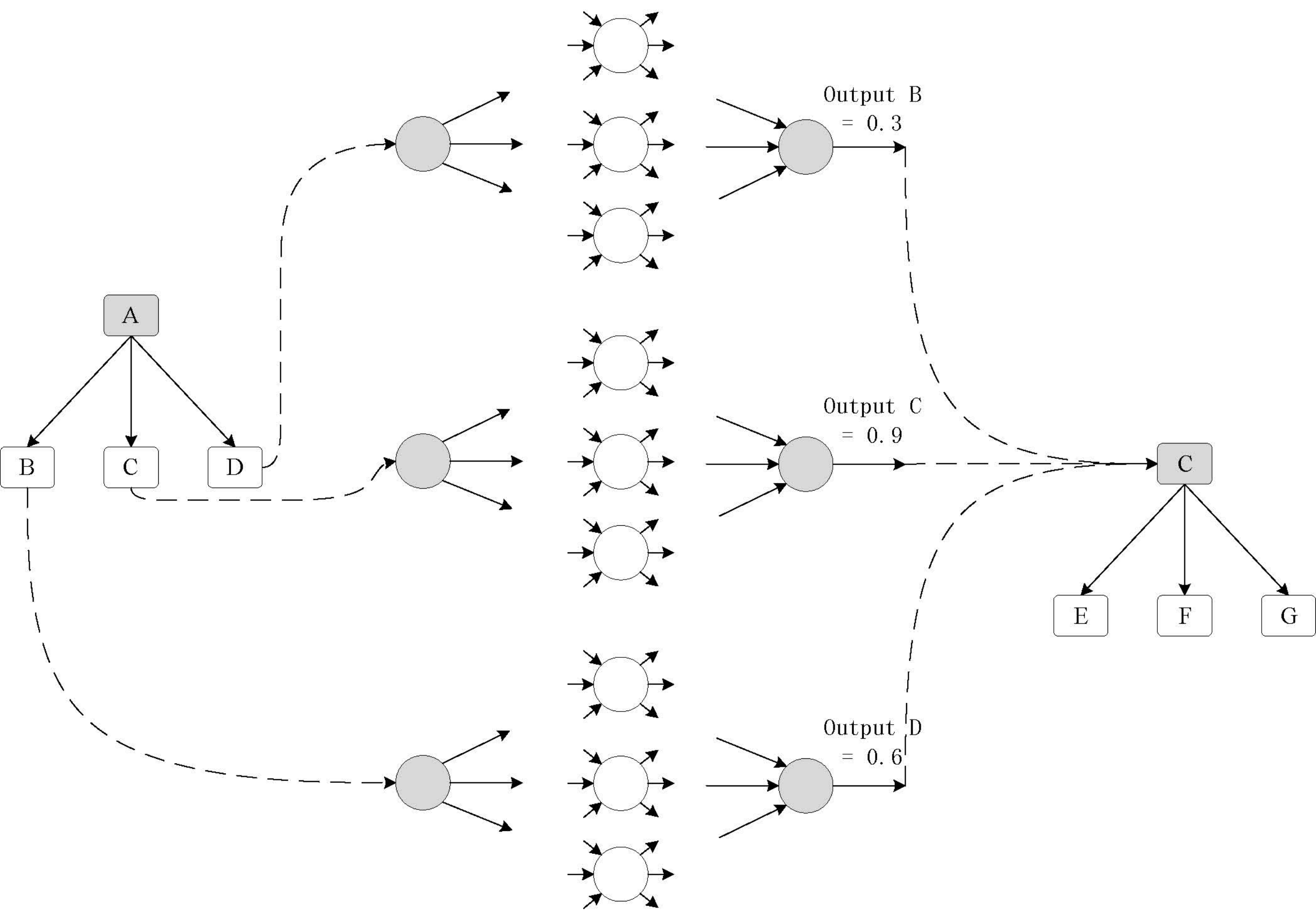}
\caption{Overview of the combination method}
\label{Fig3}       
\end{figure*}

The core idea of the method is to treat each feasible solution as an e×t matrix, and change the matrix through several predetermined change strategies, thereby gradually approaching the optimal solution. In the process of solving the problem, we use tree search to explore all the possibilities, and use DNN to decide the order to explore. Whenever a new unexplored node is found by the tree search, the DNN will predict which change strategy has the highest probability to arrive at the best solution through the predetermined change strategies, thereby determining the next search node. Fig.\ref{Fig3} shows the general flow of our method.

In this section, we first describe how to form and train a DNN model to do prediction. We then explain in detail how personnel rostering problems can be solved using tree search. Finally, we show how to use DNN in tree search to do branching decisions. \par

\subsection{Tree search}
Methods based on tree search can be used to solve optimization problems. Beginning with the root node, the search tree is explored by systematically exploring the child nodes of the root node and subsequent nodes. The solution of a given optimization problem can be understood as a leaf node in the tree.\par
In the process of tree search, each node in the tree represents a feasible arrangement for employees according to the hard constraints. The initial solution is represented by the root node. The child nodes of a node represent best solutions that can be reached by only one change from a set of predetermined change strategies.\par
\subsubsection{Change strategies}\label{c411}
There are many ways to simply modify the matrix representing a solution, such as exchanging two rows in the matrix randomly or changing some numbers in the matrix. To make the search process converge faster, some initial experiments allowed us to identify three change strategies.\par
\begin{itemize}
	\item Strategy1: randomly change 2 employees’ shifts in one day.
	\item Strategy2: randomly change 2 day’s shifts for one same employee. 
	\item Strategy3:randomly chose one shift where one of the employees doesn't need to work, and change his status to work on that shift.
\end{itemize}
\begin{figure}[htbp]
	\centering
	\includegraphics[scale=.6]{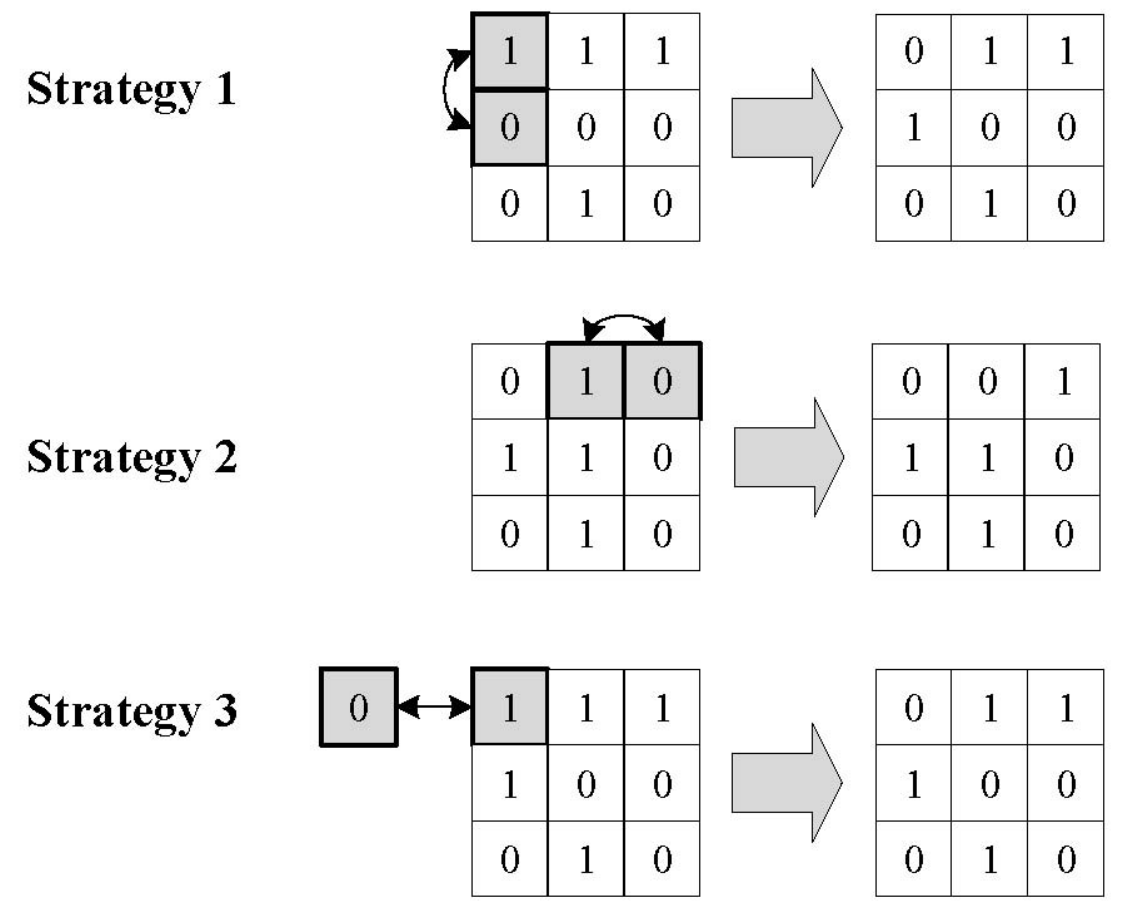}
	\caption{Three change strategies} 
	\label{Fig4}
\end{figure}
\subsection{DNN}
The DNN is a method inspired by biological neural networks. A DNN consists of multiple layers of neurons. Each neuron receives one or more weighted inputs from neurons of the previous layer, summarizes those inputs, and applies an activation function to the inputs. The value from the activation function is then sent out to the neurons of the next layer. The DNN “learns” by optimizing the weights on the arcs of the network.\citep{Hottung2020} \par
DNN can be used for both classification (the space Y consists of a set of discrete values) as well as regression (Y can take any value in R), and we use regression DNN in this work to predict the probability that a solution might be the optimal one or might become the optimal one. To substantiate the word “might”, we will use the distance to express the probability. \par
\textbf{Definition 1 (Distance)} The distance is the number of simple changes needed to be used in the current solution to become the best one.\par
\begin{figure}[htbp]
	\centering
	\includegraphics[scale=.6]{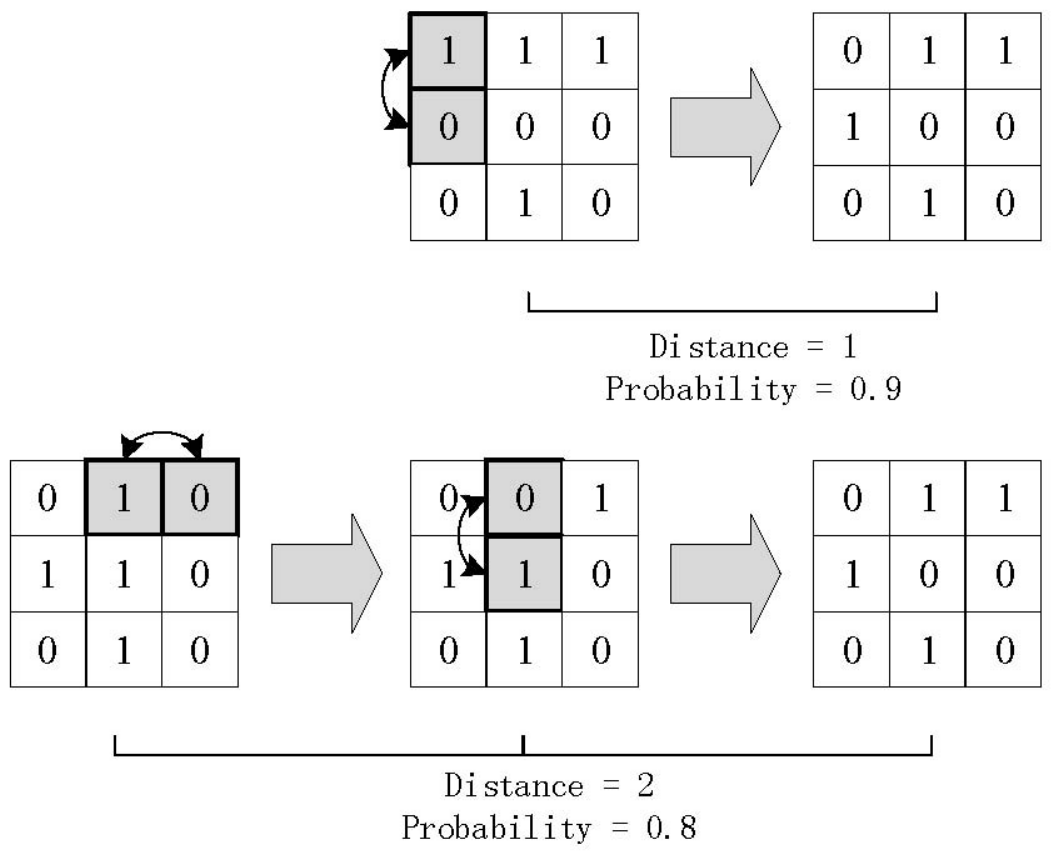}
	\caption{The link between distance and probability} 
	\label{Fig5}
\end{figure}

\textbf{Definition 2 (Probability)} The probability is used to measure the possibility that the current solution might be the optimal solution. For $distance \in N, k<0$.
\begin{equation}\label{equ11}
probability = 1-k \times distance
\end{equation}
Simple change has been explained in section \ref{c411}. Correspondingly, the shorter the distance between the current solution and the optimal solution, the higher is the probability, and the more likely the current solution might be or become the optimal solution. Fig.\ref{Fig5}  shows how to understand the distance and probability. \par

\subsubsection{DNN forming}
\begin{figure}[htbp]
	\includegraphics[scale=.6]{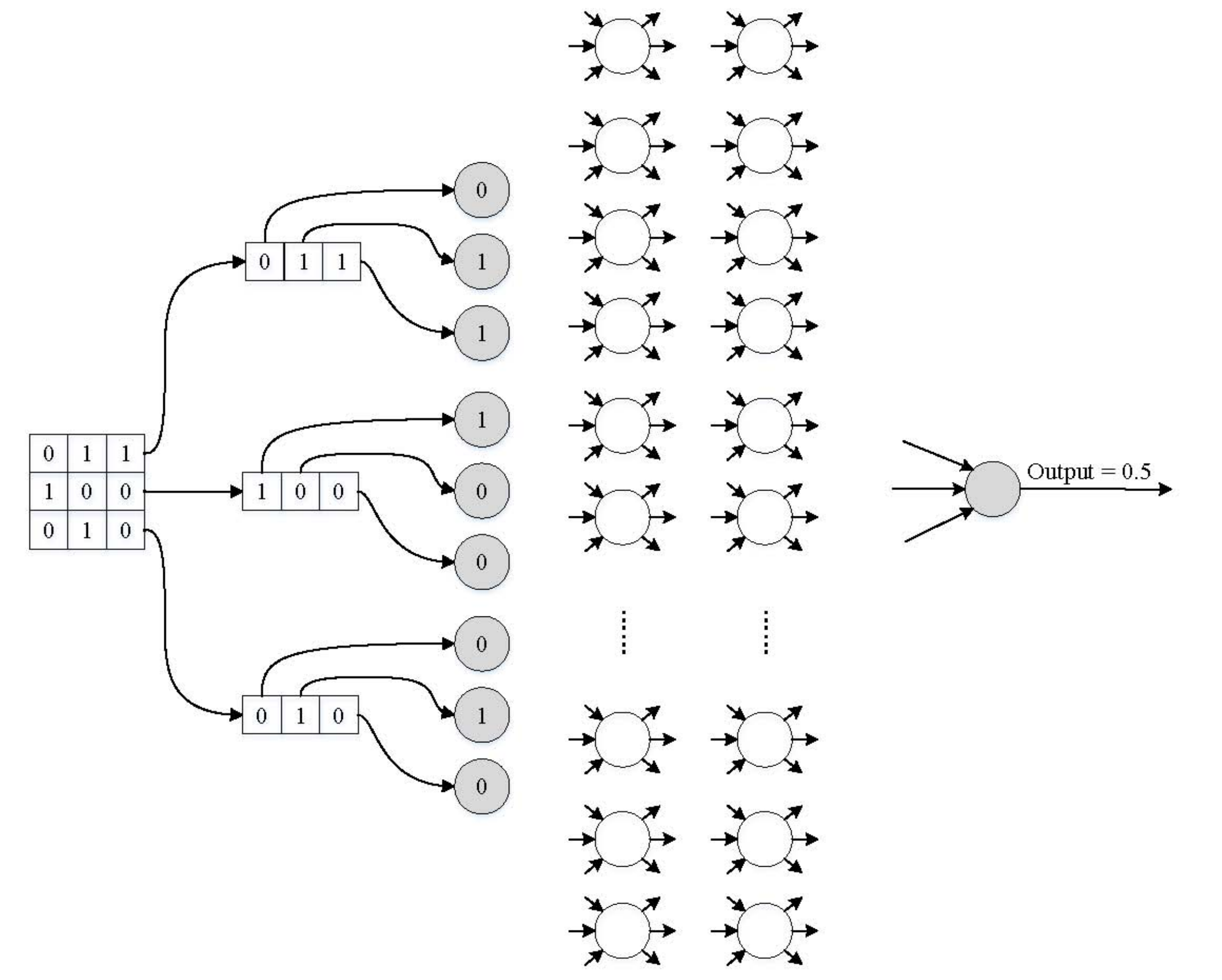}
	\caption{The matrix type input of our DNN model} 
	\label{Fig6}
\end{figure}

The DNN we use in this paper is a multi-layer DNN as Fig.\ref{Fig6} shows. The first layer is the input layer, which has $e\times t$ nodes and is used to one-dimensional the input multidimensional matrix. The middle layers are hidden layers, the number of their layers and nodes is determined through parameter adjustment. Each node in hidden layers uses the activation function, defined as $ReLU(x)=\max\left\{0,x\right\}$. The output layer contains only one node. Its activation function is a sigmoid function, defined as  $S(x)=1/(1+e^{-x})$, in order to output probability values between 0-1. As for the optimizer, we use the Adam optimizer \citep*{kingma2014adam}, which is based on a gradient descent.\par

\subsubsection{DNN training}
The work of training the DNN model is described below. As is well known, the training set is a dataset of examples used for learning, that is to fit the parameters of the model, such as weight. A set of representative instances is divided into a training set and a validation set. Also, these two kinds of sets consist of a set of data(input) and labels(output). During the process of training, each instance in the training set is inputted into the DNN, and propagated by the network, in order to generate relative output. Then these values are compared to the labels in the training data using a loss function, which is used to compute the accuracy of the prediction. In the next step, according to the influence of network weights on the loss function, the DNN model needs to adjust the weights of the network in order to reduce the value of the loss function in the next iteration (gradient descent). After dealing with all the instances in the training set, the first epoch of the training is completed. We can repeat the training for many epochs until the error doesn’t improve at all.\citep*{goodfellow2016deep}\par
As a research of solving classical problems with new methods, we have many instances that can be used as the data(input) of the training set, but there is nothing to use as the label(output) for the training set. It’s impossible to give every instance a label manually. We will use another method to obtain a relatively convincible training set for training the DNN model. The method will be explained in chapter 4.3, after the whole process is described in detail.
\subsection{DNN assisted tree search}
\subsubsection{Search Strategies}\label{c431}
There are many kinds of search strategies in tree search, such as depth-first search(DFS) of nodes that traverse the tree along the depth of the tree, breadth-first search(BFS) of nodes that traverse the tree along the width of the tree. In this paper, we refer to the idea of Depth-First-Search and improve it from different ways. New search strategies can explore the tree according to the probability value given by the DNN model and also take the penalty into consideration. We will discuss these strategies in detail below.\par
\paragraph{Depth-First-Search}
Depth-First-Search (DFS) is an algorithm for traversing or searching trees or graphs. This algorithm searches branches of the search tree as deep as possible. When the edge of node v has been searched, the search will go back to the starting node of the edge where node v was found. This process continues until all nodes have been found reachable from the source node. If there are still undiscovered nodes, select one of them as the source node and repeat the above process. The entire process is repeated until all nodes are visited. This algorithm does not adjust the execution strategy based on the information such as the structure of the graph.\par
\begin{figure}[htbp]
	\centering \includegraphics[scale=1]{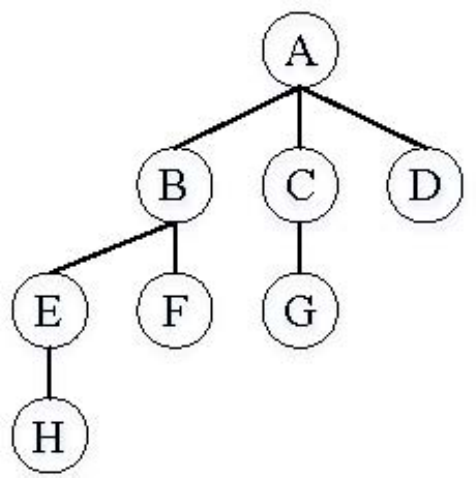}
	\caption{DFS path} 
	\label{Fig7}
\end{figure}

For example, the tree in Fig.7. According to depth-first principle, A is the initial node which is explored first. Then start from the left node of all unexplored children of A, that means B is the next node after A. The same way, after exploring B, then E rather than C, until the bottom of the branch H is explored. Next returning to an unvisited node next to B, the depth-first traversal is repeated until all nodes in the tree have been visited. The search order in Fig.\ref{Fig7} is $A\rightarrow B\rightarrow E\rightarrow H\rightarrow F\rightarrow C\rightarrow G\rightarrow D$.\par
\par
Algorithm\ref{alg1} shows the depth first search strategy. The algorithm starts with the initial solution $s_0$, stored in a node $n$, that has several properties. These are whether the current node is visited, $visited(n)$, the current penalty of associated solution, $penalty(n)$, the children nodes of the current node, $child(n)$. The penalty of the best solution $p$ and the best solution $best_n$ are set as the global value.
\begin{algorithm}[htb]
	\caption{Depth first search strategy}
	\label{alg1}
	\begin{algorithmic}
		\Require
		A node $n$ of search tree.
		\Global
		Penalty of the best solution $p$, best solution $best_n$
		\Ensure
		Node representing the best solution.
		\State \textbf{function} DNNTS-DFS($n$)
		\State \textbf{if} $isvisited(n)$ = true \textbf{then} \textbf{return} $best_n$
		\State \qquad$isvisited(n)$ = true
		\State \textbf{if} $penalty(n)<p$ \textbf{then}
		\State \qquad$p\leftarrow penalty(n)$
		\State \qquad$best_n\leftarrow n$
		\State \textbf{for} $n^{\prime}$ \textbf{in} $child(n)$
		\State \qquad DNNTS-DFS($n^{\prime}$)
		\State \textbf{return} $best_n$
	\end{algorithmic}
\end{algorithm}
\paragraph{Probability-Fist-Strategy}
This strategy also follows the principle of searching the tree as deep as possible. But we add each node in the tree a value of probability by the DNN model, and we will use it to replace the left-first order (Probability-Fist-Strategy (PFS)). That means that when exploring the child nodes of v, the child node with higher probability value will be explored first, rather than the left child node. But this principle only applies to the child nodes of the same layer and the same parent node. For nodes that are not in the same layer or nodes that are not a parent node, this principle does not apply.\par
\begin{figure}[htbp]
	\centering \includegraphics[scale=.7]{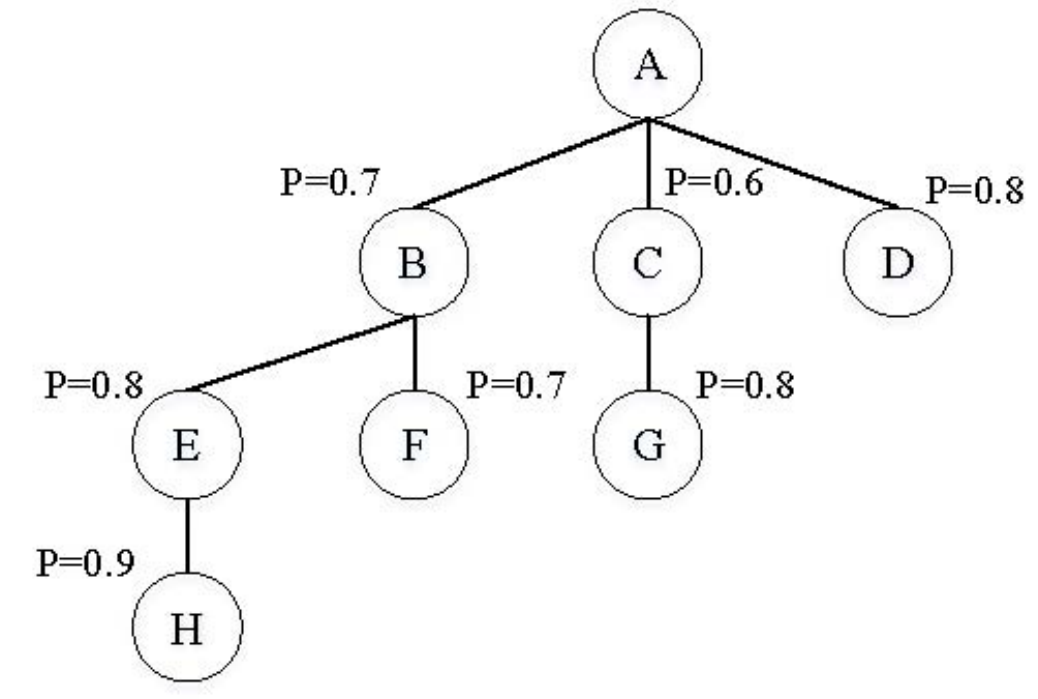}
	\caption{PFS path} 
	\label{Fig8}
\end{figure}

For example, the tree in Fig.8. node A which is explored first, but the probability of its left child node is 0.7, less than the right one. So, the next step starts from D. The same way, after exploring D, then exploring B and its subtree. The resulting search order in Fig.\ref{Fig8} is $A\rightarrow D\rightarrow B\rightarrow E\rightarrow H\rightarrow F\rightarrow C\rightarrow G$.
Algorithm\ref{alg2} shows the probability first strategy. The setting is the same as the Algorithm\ref{alg1}, but there is an additional DNN($n$), which represent the output of the DNN model when the solution associated with node $n$ is the input. i.e. The probability of solution associated with node $n$.\par
\begin{algorithm}[htb]
	\caption{Probability first strategy}
	\label{alg2}
	\begin{algorithmic}
		\Require
		A node $n$ of search tree.
		\Global
		Penalty of the best solution $p$, best solution $best_n$
		\Ensure
		Node representing the best solution.
		\State \textbf{function} DNNTS-PFS($n$)
		\State \textbf{if} $isvisited(n)$ = true \textbf{then} \textbf{return} $best_n$
		\State \qquad$isvisited(n)$ = true
		\State \textbf{if} $penalty(n)<p$ \textbf{then}
		\State \qquad$p\leftarrow penalty(n)$
		\State \qquad$best_n\leftarrow n$
		\State \textbf{Sort} $child(n)$ by DNN($n^{\prime}$) for each $n^{\prime} \in child(n)$
		\State \textbf{for} $n^{\prime}$ \textbf{in} $child(n)$
		\State \qquad DNNTS-PFS($n^{\prime}$)
		\State \textbf{return} $best_n$
	\end{algorithmic}
\end{algorithm}
\paragraph{Probability-Penalty-Strategy}
In the personnel rostering problem, the penalty determines whether the solution is optimal. When the penalty is taken into consideration is addressed as the Probability-Penalty-Strategy (PPS). We need to normalize the probability and penalty values to the same unit of measurement, and give them different weights to get a value to replace the probability value in the previous strategy.\par
\begin{algorithm}[htb]
	\caption{Probability Penalty strategy}
	\label{alg3}
	\begin{algorithmic}
		\Require
		A node $n$ of search tree.
		\Global
		Penalty of the best solution $p$, best solution $best_n$,weight for probability $w_1$, weight for penalty $w_2$
		\Ensure
		Node representing the best solution.
		\State \textbf{function} DNNTS-PPS($n$)
		\State \textbf{if} $isvisited(n)$ = true \textbf{then} \textbf{return} $best_n$
		\State \qquad$isvisited(n)$ = true
		\State \textbf{if} $penalty(n)<p$ \textbf{then}
		\State \qquad$p\leftarrow penalty(n)$
		\State \qquad$best_n\leftarrow n$
		\State \textbf{Sort} $child(n)$ by $w_1\times$ DNN($n^{\prime}$)+$w_2\times penalty(n^{\prime})$ for each $n^{\prime} \in child(n)$
		\State \textbf{for} $n^{\prime}$ \textbf{in} $child(n)$
		\State \qquad DNNTS-PPS($n^{\prime}$)
		\State \textbf{return} $best_n$
	\end{algorithmic}
\end{algorithm}
Algorithm\ref{alg3} shows the probability penalty strategy. The setting is the same as the Algorithm\ref{alg2}, but there are two additional global values $w_1$ and $w_2$, which represent the weight for probability and weight for penalty respectively.
\subsubsection{DNN assisted tree search model}
\begin{figure*}[htbp]
	\centering \includegraphics[scale=0.42]{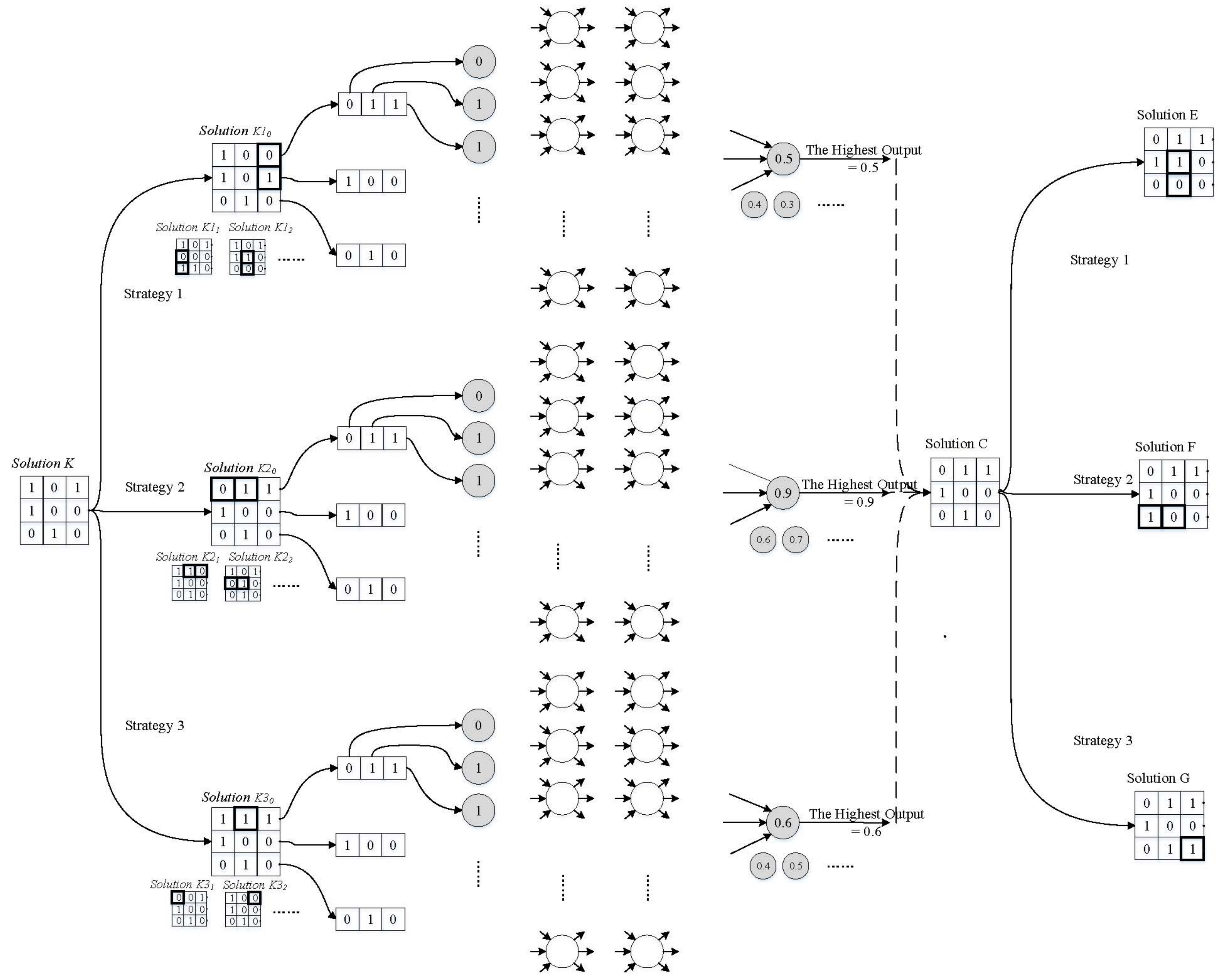}
	\caption{The process of Neural Networked Assisted Tree Search for the Personnel Rostering Problem} 
	\label{Fig9}
\end{figure*}

After forming and training the DNN model, it is used in the tree search as follows. When the node nk is to be explored, the associated $Solution K$ is changed by 3 strategies as mentioned before, iterating through all the possibilities in each strategy and get 3 sets. The set obtained through strategy 1, 2, 3 are $\left\{Solution K1_0 ,\right.$ $\left. Solution K1_1 , Solution K1_2,…\right\}$,    $\left\{Solution K2_0 , Solu- \right.$ $\left.tion K2_1 ,\right.$ $\left. Solution K2_2,…\right\}$, $\left\{Solution K3_0 , Solution -\right.$\\$\left.K3_1 , SolutionK3_2,…\right\}$. Next all possibilities are given to the DNN model. These matrices of Solution are then propagated through the DNN model. We use a $sigmoid$ activation function in the output layer to get the result between 0 and 1, this allows us to use the output as probability. Left the highest probability as the child node for each strategy, so among 3 strategies, we get 3 child nodes for each father node. The left probability is then used to decide which child node should be explored. The decision-making process is determined by the search strategies proposed in \ref{c431}, for example, exploring the node with highest probability first by PFS. Fig.\ref{Fig9} shows a whole process.\par

\subsubsection{Initial dataset}\label{c433}
We mentioned that the work applied to personnel rostering problem, although there are enough solutions (input), but no labels(output). So, we will use an automated method to generate a dataset for training the DNN model, including solution and labels.\par
First of all, for a specific personnel rostering problem that people already know the current optimal solution, we set the probability of the current optimal solution to 1, then randomly generate other solutions based on the current optimal solution. Actually, it means randomly generate $e\times t$ matrices according to the current optimal solution. And assign different labels according to the following conditions.\par
\begin{enumerate}
	\item The most important thing is to check whether the randomly generated matrix meets the hard constraint. In case the hard constraint is not satisfied, the data will not be stored in the generated data set. Only matrices meeting the hard constraint can enter the next step to judge and assign different labels.
	\item In order to give each matrix a label, we need to check the number of change strategies required(mentioned in chapter \ref{c431}) to let the current matrix change to become the optimal solution for the current specific personnel rostering problem, then use a piecewise constant function to correspond the number of changes to the label. 
	The less change strategies are needed, it means the more probable it is for the current matrix to change into the optimal solution. Equation\ref{equ12} is the piecewise constant function we use in this work.\par
\end{enumerate}
\begin{equation}\label{equ12}
f(x)=\left\{
\begin{aligned}
&0.9& \qquad 0<x\leq 3\\
&0.7& \qquad 3<x\leq 6\\
&0.5& \qquad 6<x\leq 9\\
&0.3& \qquad 9<x\leq 12\\
&0.1& \qquad x>12\\ 
\end{aligned}
\right.    	
\end{equation}\par
After generating all the data, we use these data to initially train our DNN model. After training, we can then use the trained DNN model to solve the specific personnel rostering problem which is already known current optimal solution. In a complete solution process, we can get a complete path $\left\{s_0, s_1, s_2, …,s_{op}\right\}$ from how the initial solution $s_0$ changes into the final current optimal solution sop. According to the length of the entire path and the position of the $s_k$ in the path, the distance from the current optimal solution can be found, from which the higher confidence of probability value can be given. At the same time, whenever enough data for training the DNN model is obtained, the model can be retrained. The specific problem model and initial solution $s_0$ are continuously changed to get enough data to finally complete the model training.\par
The use of these randomly generated data to train the DNN model at the beginning will not have a great impact on the efficiency of the final model, because these data are just to make the model have the ability to solve the problem, maybe the efficiency of the DNN model trained from the randomly generated data set is not very high, but at least it can ensure that the model can successfully solve a specific problem and give a complete path from the initial solution $s_0$ to the current optimal solution $s_{op}$. It is our purpose to obtain a large number of such paths, so that we can use the data and labels in these paths to retrain our DNN model with more accurate data, and through continuous iteration, our model reaches a good level.\par
\section{Result}
We now use the instance to evaluate the performance of our DNNTS method in 3 aspects. In order to measure whether our method is competitive with current methods, we experiment on a variety of personnel rostering problem instances by different methods. To find out the most efficient change strategies, we use different combinations of change strategies for comparison. We also compare DFS, PFS and PPS by the same instance for several times to find the suitable search strategy.\par
\subsection{Experimental setup} 
In order to get a good performing DNN, we need a large number of instances. We use the generator from Chapter\ref{c433} to generate 2 different types of personnel rostering problem instance sets: I1 and I2. In I1 and I2, both of them start from Monday, the horizon length of days is 14, $T = \left\{1,2,…14\right\}$. There is only 1 type of shift, the set of shifts $S = \left\{1\right\}$. 8 employees are assigned to different shifts, $E = \left\{1,2,…,8\right\}$. More detailed parameters setting are referred from Nurse Rostering Benchmark Instances\citep{Schedulingbenchmarks}.\par
We train our DNN on the instance sets mentioned above. I1 is an instance set including more than 4000 instances, which is used to train the DNN model initially. To ensure the better performance of the DNN model, we generate I2 of 2000 accurate instances, which is used to retrain the DNN model. After completing the training, we change one of the above specific problem parameters or add constraints to make it a different problem scenario, and then try to use our method to find the best solution to the new problem within a certain time limit. If the best solution cannot be found within the limited time, the current best solution is used. \par
We implement our algorithm in Python 3.7 using keras 2.3.0 under tensorflow 2.0.0 as the backend for the implementation of the DNN. All networks are trained using the Adam optimizer, which is based on a gradient descent\citep{kingma2014adam}. All experiments are conducted using Tier-2 Cluster of the Vlaams Supercomputer Center.\par
\subsection{Experiment 1: Comparison of change strategies}
The correct change strategy in the search process of the selection tree is critical to the good performance of the algorithm. Therefore, we propose three different possible combinations of change strategies for DNNTS. We train the DNN model on the I2 data set, and then we apply the DNN model to the tree search with different change strategies combinations to evaluate their impact on the performance of the algorithm.\par
The combinations of change strategies we used for comparison is as follows:\par
Combination 1:\par
\begin{itemize}
	\item Strategy1: randomly change 2 employees’ shifts in one day.
	\item Strategy2: randomly change 2 days shifts for one same employee.  
	\item Strategy3: randomly chose one shift which one of the employees who doesn’t need to work, and change his status to work on that shift.
\end{itemize}\par
Combination 2:\par
\begin{itemize}
	\item Strategy1: randomly change 2 employees’ shifts in one day.
	\item Strategy2: randomly change 2 employees’ shifts for 2 neighboring days.  
	\item Strategy3: randomly change 2 days shifts for one same employee.
\end{itemize}\par
Combination 3:\par
\begin{itemize}
	\item Strategy1: randomly change 2 employees’ shifts in one day.
	\item Strategy2: randomly change 2 employees’ shifts for 2 neighboring days.  
	\item Strategy3: randomly change 2 days shifts for one same employee.  
\end{itemize}\par
The optimization criterion is minimal penalty, the lower the better. A performance measure is the average time to solve. Another performance measure are the Figures showing declining process of the current optimal penalty. Since the final solution found by our method does not prove to be optimal, we will use final solution to represent the best solution we find in a complete algorithm process.\par
\begin{figure*}[htbp]
	\centering
	\subfigure[Penalty with the depth of the tree]{
		\begin{minipage}[t]{0.45\linewidth}
			\centering
			\includegraphics[scale=0.5]{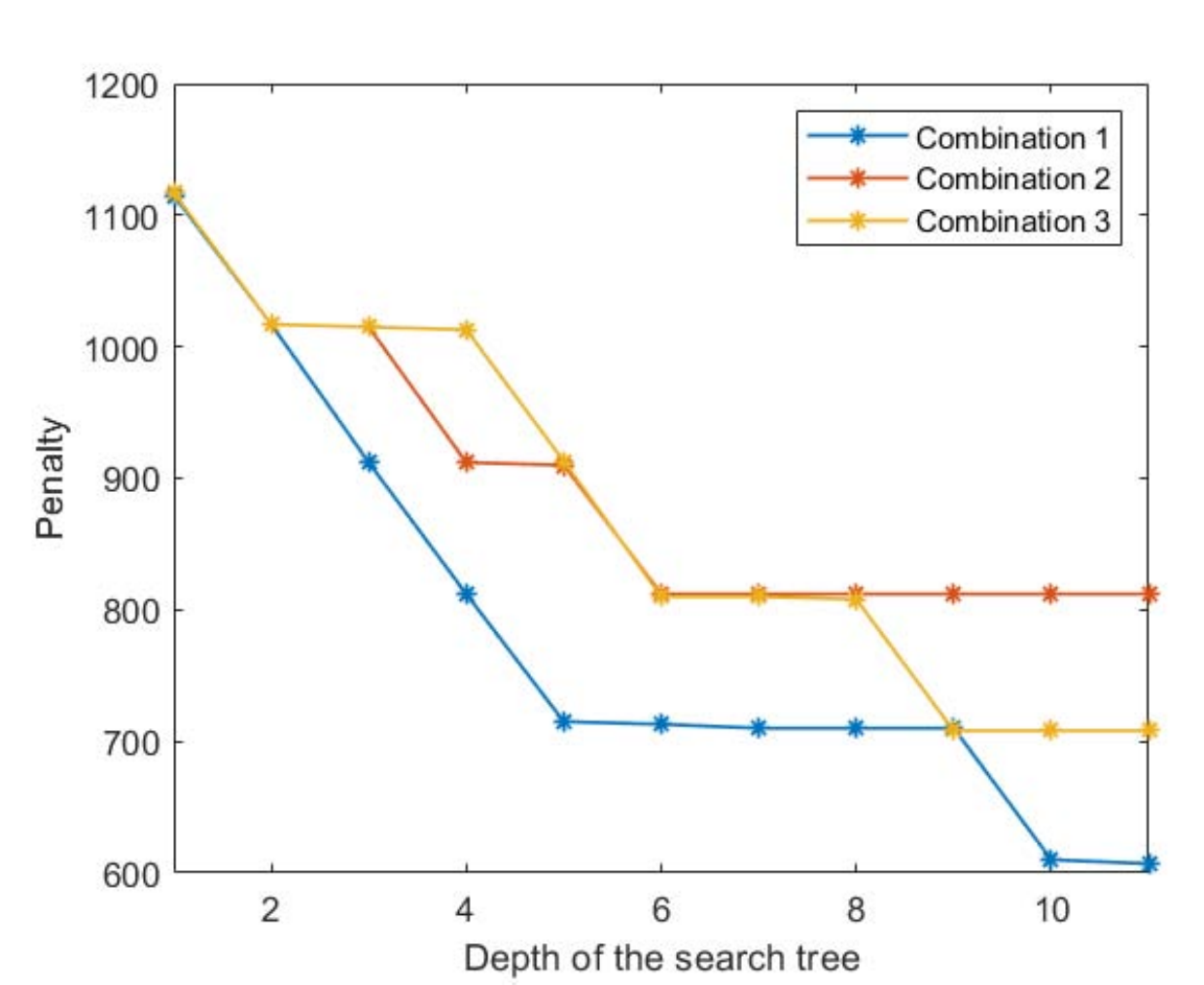}
		\end{minipage}
	}
	\subfigure[Penalty with the time]{
		\begin{minipage}[t]{0.45\linewidth}
			\centering
			\includegraphics[scale=0.5]{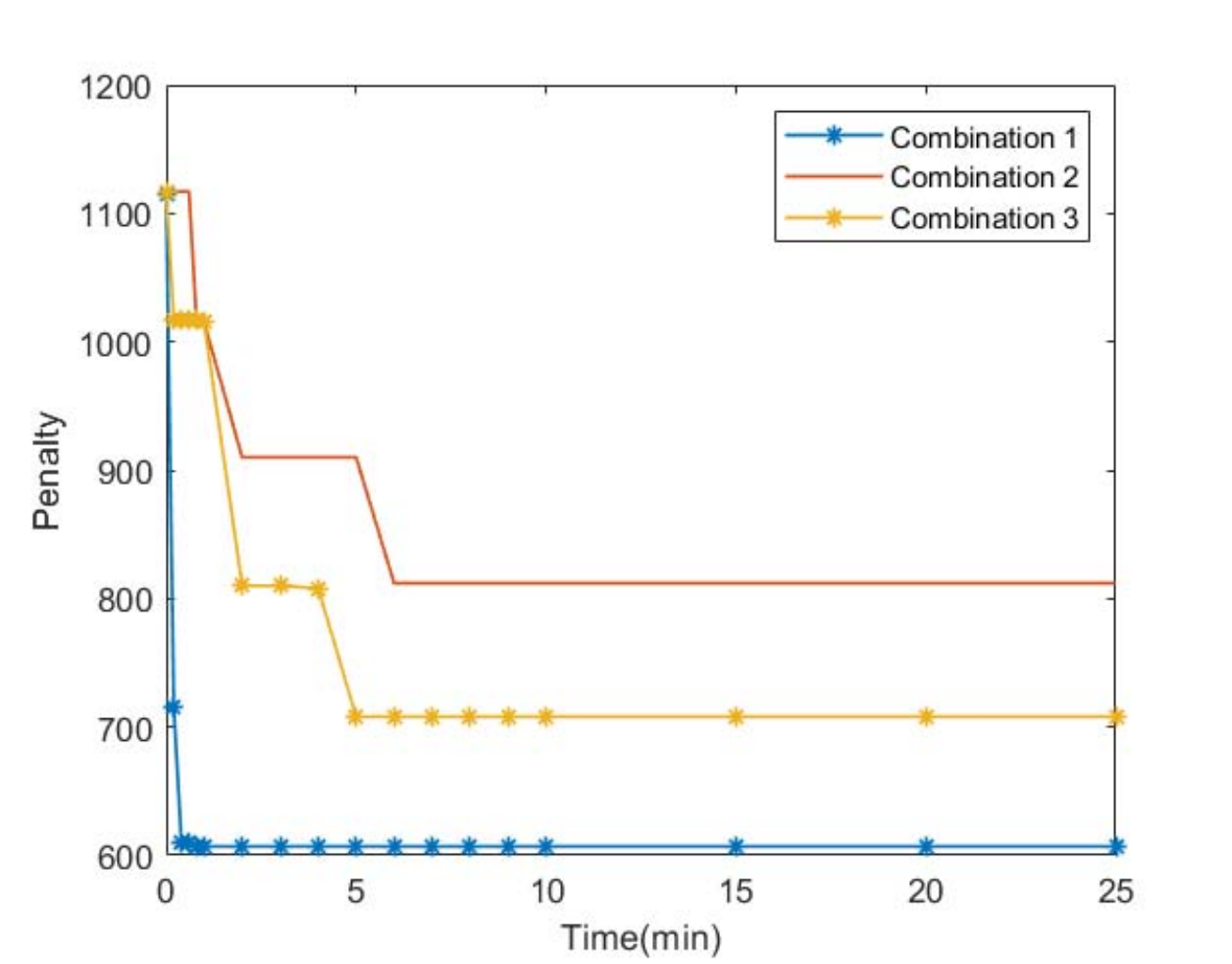}
		\end{minipage}
	}
	\centering
	\caption{The image of the convergence speed}
	\label{Fig10}
\end{figure*}

As can be seen from Table \ref{tb1}, whether the algorithm meets the stop criterion after running for 60s, 600s, or the end of the algorithm, the performance of the change strategies combination 1 is much better than the other two combinations in all aspects. As shown in the Table  \ref{tb1}, combination 1 can find the current known best solution with the penalty value of 607 in less than 1 minute. In contrast, although strategy 3 can find a relatively good solution, the total time of the entire algorithm process is too long, nearly half an hour. Although combination 2 is slightly better than combination 3 in the total time, its final solution penalty value is worse than combination 3. As can be seen from Fig.\ref{Fig10}, whether it is based on the depth of the tree search or time, the convergence speed of combination 1 is much faster than the other two combinations, and the search process will not cost too much time on these feasible solutions with similar penalty value, it almost directly comes to the final solution with the shortest path. However, the other two combinations consume too much time in the local search, which causes many platforms on the curve in Fig.\ref{Fig10}, so the convergence rate becomes slow. This is closely related to the combination of changing strategies. The strategy "Randomly change 2 employees' shifts in one day." can ensure that the elements of the solution matrix are exchanged up and down, and the strategy "Randomly change 2 employees' shifts for one same employee." can ensure that the elements of the solution matrix are exchanged left and right. The strategy "Randomly chose one shift which one of the employees who doesn't need to work, and change his status to work on that shift." can control the number of different elements in the matrix. These three strategies can guarantee that the matrix changes covers every possibility.\par

\begin{table*}[htbp]
	\centering
	\caption{Comparison of different obfuscations in terms of their transformation capabilities}
	\begin{tabular}{cccccc} 
		\toprule
		\multirow{2}{*}{Combination} & \multicolumn{2}{l}{Convergence Speed}&\quad&  \multicolumn{2}{l}{Final result}   \\
		\cline{2-3} \cline{5-6}  
		&    60s & 600s & \quad&Final penalty & Time(s) \\
		\midrule
		1 &  607 & 607  &\quad& 607 & 43.41  \\
		2 & 1015 & 812 &\quad & 812 & 680.78  \\
		3 & 1015 & 708  &\quad& 708 & 1363.25\\
		\bottomrule
	\end{tabular}
	\label{tb1}
\end{table*}

\subsection{Comparison of search strategies}
We need to compare three search strategies mentioned above in order to find the best one, namely DFS, PFS and PPS. To ensure a fair comparison among strategies, we use the DNN model trained on the I2 dataset for each strategy to evaluate the performance.  Table \ref{tb2} also provides the current best solution found in a specific period of time(60s and 600s), the final solution, and the time of the three search strategies.\par
\begin{figure*}[htbp]
	\centering
	\subfigure[Penalty with the depth of the tree]{
		\begin{minipage}[t]{0.45\linewidth}
			\centering
			\includegraphics[scale=0.5]{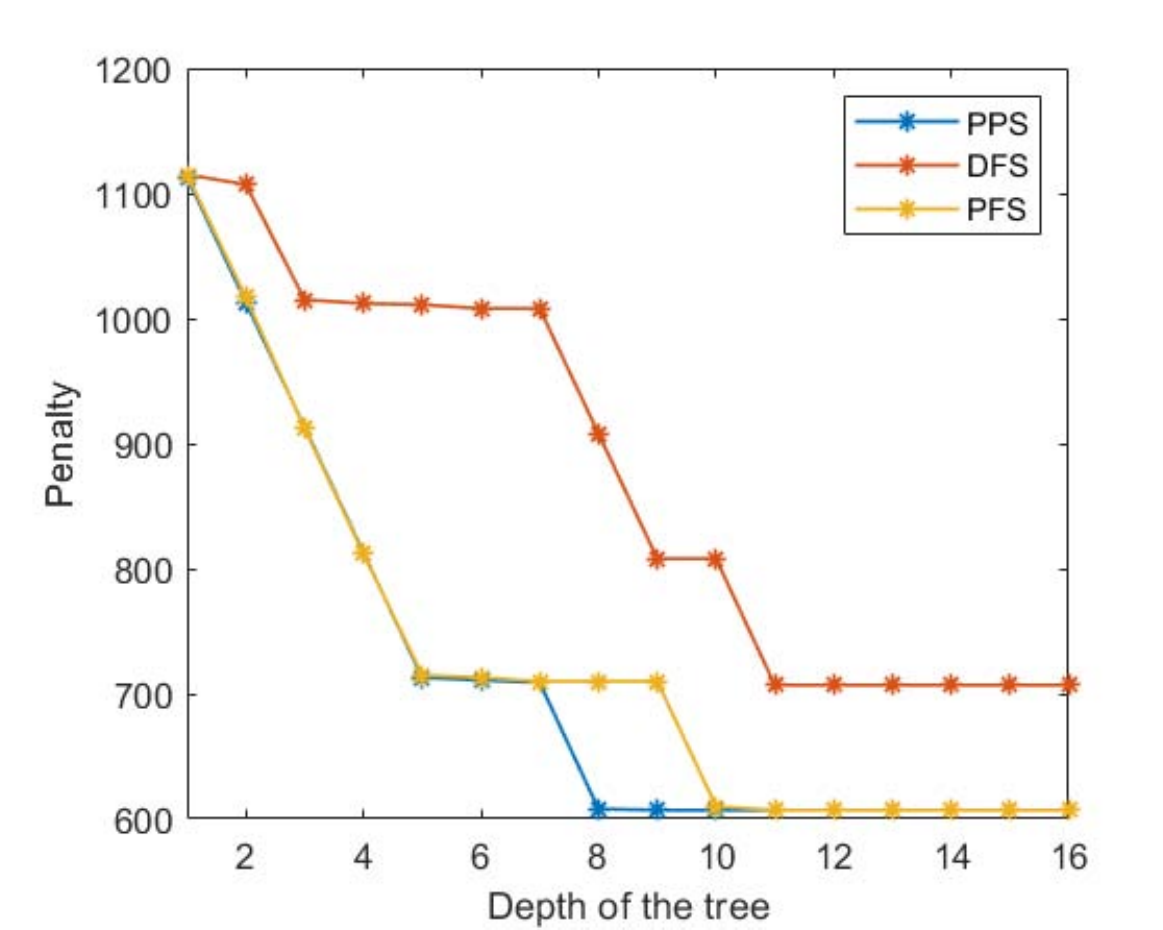}
		\end{minipage}
	}
	\subfigure[Penalty with the time]{
		\begin{minipage}[t]{0.45\linewidth}
			\centering
			\includegraphics[scale=0.5]{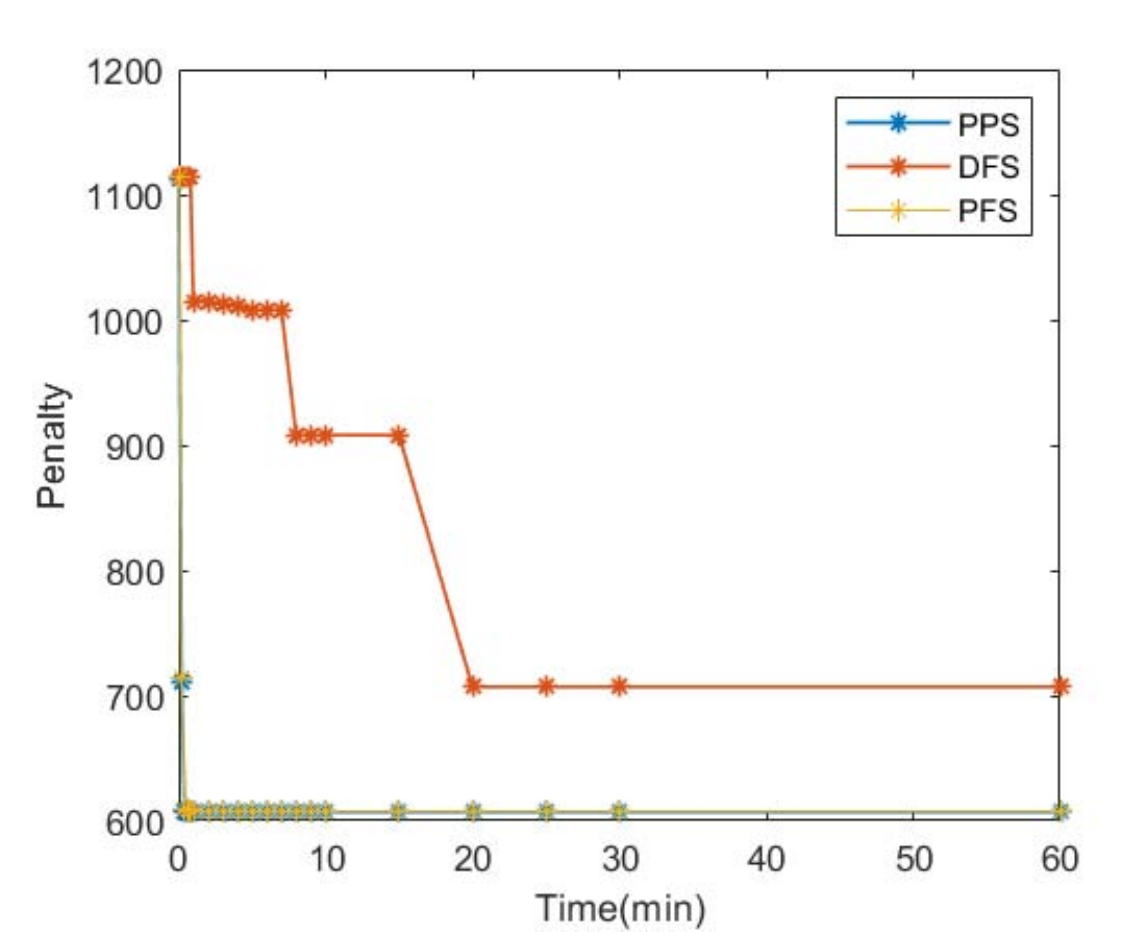}
		\end{minipage}
	}
	\centering
	\caption{The image of the convergence speed}
	\label{Fig11}
\end{figure*}

As can be seen from Table \ref{tb2}, PFS and PPS can find the final solution with the same penalty value, but PFS takes a little bit longer - just 13s. However, whether the current best solution in 60s , 600s or the final solution in the whole process, the performance of DFS is far inferior to PFS and PPS. It can be seen from Table \ref{tb2} that the introduction of the DNN model to the branch selection in the process of ordinary tree search has a certain effect. It can be further seen from Fig.\ref{Fig11} that, regardless of the depth or time of the search tree, DFS will encounter the problem that the penalty value convergence is too slow during the search process, which is actually caused by not selecting the correct branch, thus wasting more time. From the final result, the final solution of DFS stagnates at a penalty value of 707, while the convergence rate and the final solution of PPS and PFS are far superior to DFS. The final result further proves that the method of combing the DNN model with the tree search really works. Compared to PFS, the penalty value of PPS will converge slightly faster than PFS. This is why we proposed PPS. By giving Penalty a certain weight when selecting branches, it helps the search tree reach to the root as fast as possible in the beginning of the search, and the fast approach to the final solution also plays the guiding role of the DNN model.\par
\begin{table*}[htbp]
	\centering
	\caption{Comparison of three search strategies}
	\begin{tabular}{cccccc} 
		\toprule
		\multirow{2}{*}{Search Strategy} & \multicolumn{2}{l}{Convergence Speed}&\quad&  \multicolumn{2}{l}{Final result}   \\
		\cline{2-3} \cline{5-6}  
		&    60s & 600s & \quad&Final penalty & Time(s) \\
		\midrule
		DFS &  1015 & 908  &\quad& 707 & 3453.47  \\
		PFS &  607 & 607  &\quad& 607 & 43.41  \\
		PPS & 607 & 607  &\quad& 607 & 30.54\\
		\bottomrule
	\end{tabular}
	\label{tb2}
\end{table*}
Compared with ordinary search methods, because of the guidance of the DNN model, the number of nodes searched by DNNTS is many orders of magnitude less. Usually the best results can be achieved in twenty selections, which clearly shows that the DNN model is very effective.\par

\par
\subsection{Experiment 3: Evaluation of methods}
We compare the DNNTS and Roster viewer \citep{curtois2014computational}, which has two built-in solutions-VDS (3.11) and Branch and Price (B$\&$P). We compare these three methods on eight Groups with different problem types and report the performance of each method. Group 1 and Group 2 are designed by ourselves, and Group 3-11 are the instances from Nurse Rostering Benchmark Instances\citep{Schedulingbenchmarks}. Method VDS requires a limit on the maximum time. So we first run DNNTS, and set the maximum time of VDS equal to the time costed by DNNTS. So that we can better observe the performance of the different methods during the same time. The parameters of different group are shown in Table \ref{tb3}. \par   
\begin{table*}[htbp]
	\centering 
	\caption{\label{tb3}Parameters of each Group} 
		\begin{tabular}{cccccc} 
			\toprule 
			Group&Weeks&Employees&Shift types&Hard constraints&Best known Solution\\ 
			\midrule 
			1&2&8&1&2&/\\
			2&2&8&1&3&/\\
			3&2&8&1&10&607\\
			4&2&14&2&10&828\\
			5&2&20&3&10&1001\\
			6&4&10&2&10&1716\\
			7&4&16&2&10&1143\\
			8&4&18&3&10&1950\\
			9&4&20&3&10&1056\\
			10&4&30&4&10&1300\\
			11&6&45&6&10&3833\\
			\bottomrule 
	\end{tabular} 
\end{table*}

The results are shown in Table \ref{tb4}. For small instances (Group 1 to Group 9), B$\&$P can find the best solution in a really short time, but for big instances(Group 10 and Group 11), it meets some problem. This is the exponential explosion problem often encountered in B$\&$P method. Since there is a DNN model to do branch decision, our DNNTS method consumes acceptable time in all groups. In terms of the quality of the final solution, The performance of VDS isn't good, probably because it is an traversal algorithm, so maybe it needs more time to get better results. Some penalty value of the final solution found by DNNTS are the same as the best known solutions, others are near to the best known solutions, which proves our method can be applied to the public instance set and propagated. The result shows that DNNTS can compete with the latest methods in terms of solution time and solution quality. And the result also illustrates the positive significance of introducing the DNN model into the tree search, and also emphasizes the importance of adequate training of the DNN model.\par
\begin{table*}[htbp]
	\centering
	\caption{Comparison to other methods on the test set}
		\begin{tabular}{ccccccccc} 
			\toprule
			\multirow{2}{*}{Group} &\multirow{2}{*}{Best Known Penalty}&  \multicolumn{3}{l}{Final Penalty}&\quad&  \multicolumn{3}{l}{Time(s)}   \\
			\cline{3-5} \cline{7-9} 
			&& DNNTS& B$\&$P &VDS& \quad&DNNTS & B$\&$P&VDS \\
			\midrule
			1 & / & 0 &0 & 0 &\quad& 1.98 & 0  & 2 \\
			2 & / & 1 & 1 &1 &\quad& 1.60 &  0 & 2\\
			3 & 607 & 607 & 607 & 607 &\quad& 43.42 & 0.27 & 44 \\
			4 & 828 & 828 & 828 & 937 &\quad& 205.47 & 0.13 & 206 \\
			5 & 1001 & 1001 & 1001 & 1103 &\quad& 170.77 & 0.45 & 171\\
			6 & 1716 & 1718 & 1716 & 1721 &\quad& 283.9 & 1.5 & 284\\
			7 & 1143 & 1143 & 1160 & 1636 &\quad& 78.28 & 25.61 & 79\\
			8 & 1950 & 1952 & 1952 & 2340 &\quad& 1800.78 & 10.45 & 1801\\
			9 & 1056 & 1057 & 1058 & 1278 &\quad& 1589.32 & 93.73 &1590 \\
			10 & 1300 & 1317 & 1308 & 2643 &\quad& 560.93 & 11831.06 & 561\\
			11 & 3827 & 4378 & / & 5514 &\quad& 2660.91 & Out of memory & 2661\\
			\bottomrule
	\end{tabular}
	\label{tb4}
\end{table*}

\section{Conclusion and future work}
We propose a method aim to the personnel rostering problem combining DNN and tree search, which uses deep learning to assist branch selection. We prove that compared with the examples in the literature, DNNTS finds good solutions equal or near to the best known solutions, which are able to compare with the rostering problem solver. DNNTS can solve problems with very little user input. It mainly relies on the current optimal solution provided to learn how to build a solution by itself. There are many ways for DNNTS to work in the future. If we can model other optimization problem as a standard model with the following characteristics, applying DNNTS to other similar optimization problems is a possible choice.\par
\begin{itemize}
	\item The solutions of the problems can be regarded as m×n matrices, so that they can be transferred as the input of the neural network.
	\item The best solution can be found by some changes from the initial solution, so that some child nodes can be derived at each parent node. 
	\item The problems have some soft constraints, so that we can use the penalty to set the lower bound of the tree search.
\end{itemize}
Other areas of future work include the use of reinforcement learning and others to further improve efficiency. We suggest that by using a faster programming language or using a GPU instead of a neural network's CPU, the runtime of the results we obtained may be improved. In addition, many changes can be made to DNNTS, such as reconfiguring the DNN network structure or adjusting branch pruning functions. These changes can improve performance in terms of runtime and solution quality.\par

\bibliographystyle{nameyear}

\bibliography{ref}

%
%

\end{document}